\documentclass[10pt,twocolumn,letterpaper]{article}

\usepackage{cvpr}              
\makeatletter
\@namedef{ver@everyshi.sty}{}
\makeatother
\usepackage{tikz}
\usepackage{graphicx}
\usepackage{amssymb}
\usepackage{booktabs}
\usepackage{algorithmic}
\usepackage[ruled]{algorithm2e}
\usepackage{color}
\usepackage{xspace}
\usepackage{xcolor}
\usepackage{tikz}
\usepackage{comment}
\usepackage{mathrsfs}
\usepackage{amsmath}
\usepackage{diagbox}
\usepackage{booktabs}
\usepackage{multirow}
\usepackage{multicol}
\usepackage{bm}
\usepackage{amsfonts}
\usepackage{cancel}
\usepackage{tabularx}
\usepackage{color}
\usepackage{xcolor,colortbl}
\usepackage[inline]{enumitem}
\usepackage{pifont}
\usepackage{xspace}
\usepackage[accsupp]{axessibility}
\newcommand{\system}{ResFormer\xspace}
\newcommand{\ra}[1]{\renewcommand{\arraystretch}{#1}}

\newcommand{\xmark}{\ding{55}}

\definecolor{citecolor}{RGB}{0, 113, 188}
\definecolor{risecolor}{RGB}{0, 139, 139}
\definecolor{dropcolor}{RGB}{119, 136, 153}

\newcommand{\rise}[1]{\textcolor{risecolor}{\scriptsize{($\uparrow$#1)}}}
\newcommand{\drop}[1]{\textcolor{dropcolor}{\scriptsize{($\downarrow$#1)}}}
\newcommand{\equal}[1]{\textcolor{dropcolor}{\scriptsize{($\leftrightarrow$#1)}}}

\definecolor{Gray}{gray}{0.9}
\newcolumntype{a}{>{\columncolor{Gray}}c}

\usepackage[pagebackref=true,breaklinks=true,letterpaper=true,colorlinks,citecolor=citecolor, bookmarks=false]{hyperref}

\usepackage[capitalize]{cleveref}
\crefname{section}{Sec.}{Secs.}
\Crefname{section}{Section}{Sections}
\Crefname{table}{Table}{Tables}
\crefname{table}{Tab.}{Tabs.}

\newcommand\blfootnote[1]{%
  \begingroup
  \renewcommand\thefootnote{}\footnote{#1}%
  \addtocounter{footnote}{-1}%
  \endgroup
}

\def\cvprPaperID{***} 
\def\confName{CVPR}
\def\confYear{2023}
\begin{document}

\title{ResFormer: Scaling ViTs with Multi-Resolution Training}

\author{Rui Tian$^{1,2}$ \ \ \ \ 
Zuxuan Wu$^{1,2\dagger}$ \ \ \ \ 
Qi Dai$^3$ \ \ \ \ 
Han Hu$^3$ \ \ \ \ 
Yu Qiao$^4$ \ \ \ \ 
Yu-Gang Jiang$^{1,2}$ \vspace{0.1in}\\
{$^1$Shanghai Key Lab of Intell. Info. Processing, School of CS, Fudan University} \\
{$^2$Shanghai Collaborative Innovation Center of Intelligent Visual Computing} \\
{$^3$Microsoft Research Asia}  \quad 
{$^4$Shanghai AI Laboratory} 
}
\maketitle

\begin{abstract}
Vision Transformers (ViTs) have achieved overwhelming success, yet they suffer from vulnerable resolution scalability, \ie, the performance drops drastically when presented with input resolutions that are unseen during training. We introduce, \system, a framework that is built upon the seminal idea of multi-resolution training for improved performance on a wide spectrum of, mostly unseen, testing resolutions. In particular, \system operates on replicated images of different resolutions and enforces a scale consistency loss to engage interactive information across different scales. More importantly, to alternate among varying resolutions effectively, especially novel ones in testing, we propose a global-local positional embedding strategy that changes smoothly conditioned on input sizes. We conduct extensive experiments for image classification on ImageNet. The results provide strong quantitative evidence that \system has promising scaling abilities towards a wide range of resolutions. For instance, \system-B-MR achieves a Top-1 accuracy of 75.86\% and 81.72\% when evaluated on relatively low and high resolutions respectively (\ie, 96 and 640), which are 48\% and 7.49\% better than DeiT-B. We also demonstrate, moreover, \system is flexible and can be easily extended to semantic segmentation, object detection and video action recognition. Code is available at \href{https://github.com/ruitian12/resformer}{https://github.com/ruitian12/resformer}.
\blfootnote{$^{\dagger}$Corresponding author.}
\blfootnote{Note that we use resolution, scale and size interchangeably.}

\end{abstract}

\section{Introduction}
\label{sec:intro}
The strong track record of Transformers in a multitude of Natural Language Processing~\cite{vaswani2017attention} tasks has motivated an extensive exploration of Transformers in the computer vision community. At its core, Vision Transformers (ViTs) build upon the multi-head self-attention mechanisms for feature learning through partitioning input images into patches of identical sizes and processing them as sequences for dependency modeling. Owing to 
their strong capabilities in capturing relationships among patches, ViTs and their variants demonstrate prominent results in versatile visual tasks, \eg, image classification~\cite{liu2021swin,touvron2021training,yuan2021tokens,zhanghivit}, object detection~\cite{carion2020end,li2022exploring,wang2021pyramid}, vision-language modeling~\cite{radford2021learning,jia2021scaling,wang2022omnivl} and video recognition~\cite{bertasius2021space,li2022uniformer,liu2022video,xing2023svformer}.

While ViTs have been shown effective, it remains unclear how to scale ViTs to deal with inputs with varying sizes for different applications. For instance, in image classification, the \textit{de facto} training resolution of 224 is commonly adopted~\cite{liu2021swin,touvron2021training,touvron2022deit,yuan2021tokens}. However, among works in pursuit of reducing the computational cost of ViTs~\cite{meng2022adavit,rao2021dynamicvit}, shrinking the spatial dimension of inputs is a popular strategy~\cite{wang2021not,lin2022super,chen2022coarse}. On the other hand, fine-tuning with higher resolutions (\eg, 384) is widely used~\cite{dosovitskiy2020image,liu2021swin,touvron2022deit,steiner2022how,wu2021cvt,xie2022simmim} to produce better results. Similarly, dense prediction tasks such as semantic segmentation and object detection also require relatively high resolution inputs~\cite{ali2021xcit,liu2022swin,wang2021pyramid,li2022exploring}.

\begin{figure}[t]
  \centering
    \includegraphics[width=0.95\linewidth]{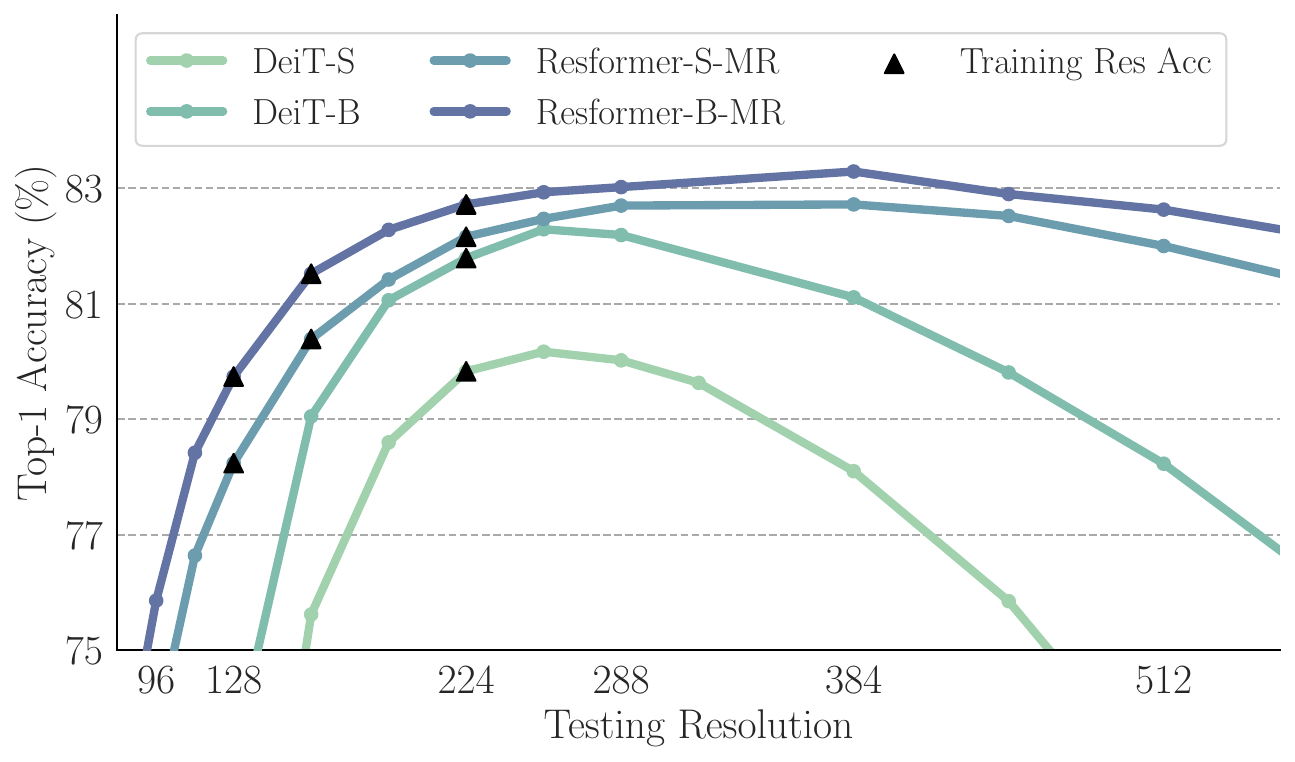}
    \vspace{-0.12in}
   \caption{Comparisons between \system and vanilla ViTs. \system achieves promising results on a wide range of resolutions.}
   \label{fig:intro_res}
\vspace{-0.1in}
\end{figure}

Despite of the necessity for both low and high resolutions, limited effort has been made to equip ViTs with the ability to handle different input resolutions. Given a novel resolution that is different from that used during training, a common practice adopted for inference is to keep the patch size fixed and then perform bicubic interpolation on positional embeddings directly to the corresponding scale. As shown in \cref{sec:explore_res}, while such a strategy is able to scale ViTs to relatively larger input sizes, the results on low resolutions plunge sharply. In addition, significant changes between training and testing scales also lead to limited results (\eg, DeiT-S trained on a resolution of 224 degrades by 1.73\% and 7.2\% when tested on 384 and 512 respectively).

Multi-resolution training, which randomly resizes images to different resolutions, is a promising way to accommodate varying resolutions at test time. While it has been widely used by CNNs for segmentation~\cite{he2017mask},  detection~\cite{he2016deep} and action recognition~\cite{wu2020multigrid}, generalizing such an idea to ViTs is challenging and less explored. For CNNs, thanks to the stacked convolution design, all input images, regardless of their resolutions, share the same set of parameters in multi-resolution training. For ViTs, although it is feasible to share parameters for all samples, bicubic interpolations of positional embeddings, which are not scale-friendly, are still needed when iterating over images of different sizes.  

In this paper, we posit that positional embeddings of ViTs should be adjusted smoothly across different scales for multi-resolution training. The resulting model then has the potential to scale to different resolutions during inference. Furthermore, as images in different scales contain objects of different sizes, we propose to explore useful information across different resolutions for improved performance in a similar spirit to feature pyramids, which are widely used in hierarchical backbone designs for both image classification~\cite{liu2021swin,he2016deep} and dense prediction tasks~\cite{lin2017feature,he2017mask,he2015spatial}.

To this end, we introduce \system, which which takes in inputs as  multi-resolution images during training and explores multi-scale clues for better results. Trained in a single run, \system is expected to generalize to a large span of testing resolutions. In particular, given an image during training, \system resize it to different scales, and then use all scales in the same feed-forward process. To encourage information interaction among different resolutions, we introduce a scale consistency loss, which bridges the gap between low-resolution and high-resolution features by self-knowledge distillation. More importantly, to facilitate multi-resolution training, we propose a global-local positional embedding strategy, which enforces parameter sharing and changes smoothly across different resolutions with the help of convolutions. Given a novel resolution at testing, \system dynamically generates a new set of positional embeddings and performs inference.

To validate the efficacy of \system, we conduct comprehensive experiments on ImageNet-1K~\cite{deng2009imagenet}. We observe that \system makes remarkable gains compared with vanilla ViTs which are trained on single resolution. Given the testing resolution of 224, \system-S-MR trained on resolutions of 128, 160 and 224 achieves a Top-1 accuracy of 82.16\%, outperforming the 224-trained DeiT-S~\cite{touvron2021training} by 2.24\% . More importantly, as illustrated in \cref{fig:intro_res}, \system surpasses DeiT by a large margin on unseen resolutions, \eg, \system-S-MR outperforms DeiT-S by 6.67\% and 56.04\% when tested on 448 and 80 respectively. Furthermore, we also validate the scalability of \system on dense prediction tasks, \eg, \system-B-MR achieves 48.30 mIoU on ADE20K~\cite{zhou2019semantic} and 47.6 AP$^{\text{box}}$ on COCO~\cite{lin2014microsoft}. 
We also show that \system can be readily adapted for video action recognition with different sizes of inputs via building upon TimeSFormer~\cite{bertasius2021space}.

\section{Related Work}
\noindent \textbf{Scaling Vision Models.} 
Many studies in recent literature \cite{tan2019efficientnet,riquelme2021scaling,liu2022swin,zhai2022scaling} discuss how to scale vision models, with most of them focusing on the capacity of deep neural networks.
For instance, EfficientNet \cite{tan2019efficientnet} studies how model width, depth and input resolution affect convolutional neural networks.
RegNet~\cite{radosavovic2020designing} designs manual designing space for CNNs and finds simple linear correlation between the search space (\emph{e.g.} width) and performance.
ResNet-RS~\cite{bello2021revisiting} presents how different scaling strategies on depth and input resolution can affect the model capacity.

Recent approaches have investigated scaling of transformers~\cite{liu2022swin,zhai2022scaling}. For example, V-Moe~\cite{riquelme2021scaling} scales vision transformers to large model sizes with sparse mixture-of-experts. Several studies~\cite{kolesnikov2020big,goyal2021self,el2021large,xie2022data} explore the aspect of data scaling under self-supervised framework, \ie, how data sizes affect the model performance. In contrast, limited effort has been made towards the scaling abilities of models towards input resolutions. Attempts have been made by Liu \etal~\cite{liu2022swin} to scale up to larger resolutions, while neglecting lower resolutions. Instead, our work takes the initiative to scale ViTs to various resolutions, both lower and higher, satisfying the practical needs from varied visual tasks.

\vspace{0.05in}
\noindent \textbf{Positional embedding.} The self-attention architecture is clueless about spatial relationships among patches. Therefore, to overcome permutation-invariance, various positional embedding strategies have been proposed to enable Transformers to perceive the sequence order of input tokens. Absolute positional embeddings (APE) infuse global spatial information into Transformers, \eg, sine-cosine APE~\cite{vaswani2017attention} proposed for NLP tasks and learned APE adopted in the vanilla Vision Transformer~\cite{dosovitskiy2020image}. Meanwhile, efficacy of relative position embeddings (RPB) is widely validated in both language~\cite{dai2019transformer,shaw2018self} and vision tasks~\cite{liu2022swin,wu2021rethinking,chen2021regionvit}. 
For instance, Wu \etal replaces APE with the relative strategy of iRPE~\cite{wu2021rethinking} for performance gains on classification and detection. 
ConViT~\cite{d2021convit} also suggests that adding gated relative positional embeddings to self-attention blocks brings about soft convolutional inductive biases. Moreover, dynamic positional embeddings are introduced to model local information from input tokens, \eg, Twins~\cite{chu2021twins} adopt conditional positional embeddings (CPE)~\cite{chu2021conditional} implemented by convolutions. In order to improve performance and scalability simultaneously, we propose to inject spatial embeddings in \system from both from global and local perspectives.

\vspace{0.05in}
\noindent \textbf{Multi-scale training.} In early CNNs, multi-scale data augmentations~\cite{simonyan2014very} are employed for image classification by randomly sampling training images from a certain range of scales. Later in dense prediction tasks, multi-scale training and testing become an widely-adopted paradigm~\cite{carion2020end,singh2018analysis,law2018corner}. In addition, the idea has also been explored in action recognition. Wu \etal introduce a multigrid strategy~\cite{wu2020multigrid}, which enables efficient training by sampling data with different grids of temporal span, spatial span and temporal stride.  Video ResKD~\cite{ma2022rethinking} achieves excellent efficiency by employing high-resolution features from large models as teachers to improve low-resolution performance. 

Most approaches in multi-scale training rely on CNNs, as convolutions can be readily applied to varying sizes of inputs.
In contrast, vanilla ViTs are equipped with tokens of fixed-dimension, other related attempts lay emphasis on multi-scale spatial dimension of features instead of input~\cite{gu2022multi,fan2021multiscale} or perform in an unsupervised way~\cite{ranasinghe2022self}, yet limited effort has been made to explore multi-resolution supervised training for ViTs. In this paper, we make the first step to investigate such a strategy for ViTs which not only leads to good performance on training resolutions but can also generalize towards novel resolutions.

\section{Resolution Generalization} \label{sec:explore_res}
In this section, we conduct a set of pilot experiments to show the scalability of ViTs towards different resolutions.

\begin{figure}[t]
\centering
\includegraphics[width=0.95\linewidth]{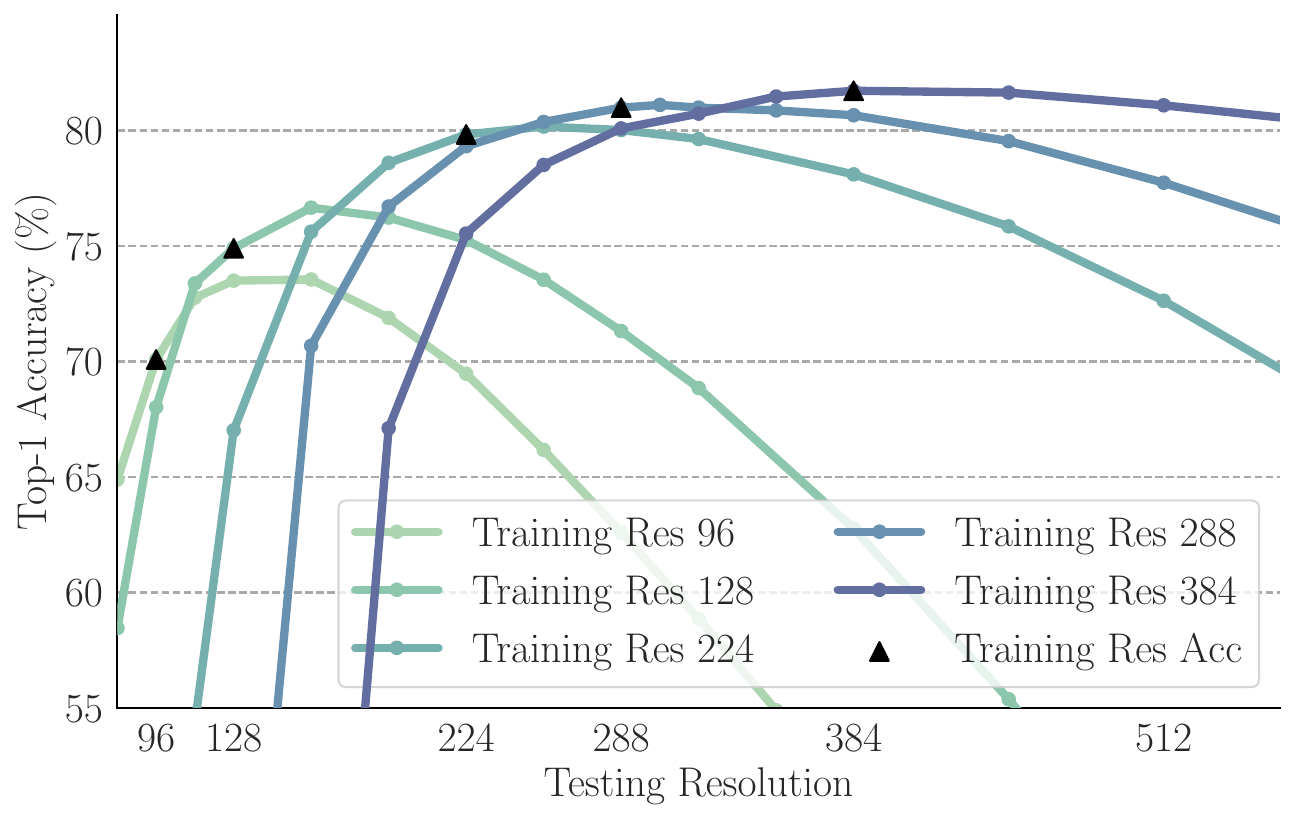}
\vspace{-0.1in}
\caption{Top-1 accuracy of DeiT-S trained with 5 different resolutions and tested on resolutions varying from $80$ to $576$. During testing, we follow the common  pre-processing steps (\ie, $\texttt{Resize}$ and $\texttt{CenterCrop}$ in Pytorch implementation) and set the cropping rate to 0.875. }
\vspace{-0.1in}
\label{fig:res_map}
\end{figure}

\vspace{0.05in}
\noindent \textbf{Generalizing to different resolutions.} As revealed by previous work~\cite{touvron2019fixing}, CNNs suffer from distribution shifts between training and testing due to different pre-processing methods, \ie, the \textit{de facto} ``random resizing and cropping'' strategy for training and ``center cropping'' for testing result in different distributions of cropped regions in images. For ViTs, theoretically, the discrepancy persists since the same pre-processing strategies of training and testing are employed. However, there lacks a comprehensive study on how ViTs behave towards input scales varied from the training process. To this end, we feed pre-trained ViTs with testing samples of varying sizes. In particular, we instantiate ViT models with DeiT-S~\cite{touvron2021training} and initialize the model with weights pre-trained on ImageNet-1K. We then fine-tune the model on a resolution of 96, 128, 224, 288 and 384 respectively.
~\footnote{Please refer to \cref{sec:appendix_image} for detailed fine-tuning setup.}
These derived models are then tested on a broad spectrum of resolutions. Following~\cite{touvron2021training,liu2021swin}, we simply resize the position embeddings with bicubic interpolation on different testing resolutions. The results are shown in \cref{fig:res_map}. We observe the following trends for scaling up or down:
\vspace{-0.05in}

\begin{itemize}[leftmargin=*]
 \setlength\itemsep{0.01em}
\item Scaling down: All models undergo severe performance drop when directly adapted to small-scale inputs, especially for ones pre-trained on larger resolutions. For example, the Top-1 accuracy of DeiT-S with a training resolution of 384 decreases by 6.18\% when tested on 224 and even plummets below 30\% when the testing resolution is further reduced to 160.  
\item Scaling up: Ideally, increasing testing resolutions results in improved accuracy, which is also suggested as byproduct of train-test distribution discrepancy in~\cite{touvron2019fixing}. However, models yield unsatisfactory performance when gap enlarges. \eg, DeiT-S with a low training resolution of 128 stops growing in accuracy when the testing resolution reaches 256. It achieves a Top-1 accuracy of 75.26\% with testing resolution set to 224, which is 4.57\% lower than model trained with a resolution of 224, directly. 
\end{itemize}

\vspace{-0.05in}
Above all, ViTs are vulnerable to resolution discrepancies between training and testing, particularly when evaluated on low-resolution inputs. This motivates us to equip ViTs with scalability towards a wide range of test resolutions so as to meet the need of versatile applications.

\begin{figure*}[t]
\centering
\includegraphics[width=\linewidth]{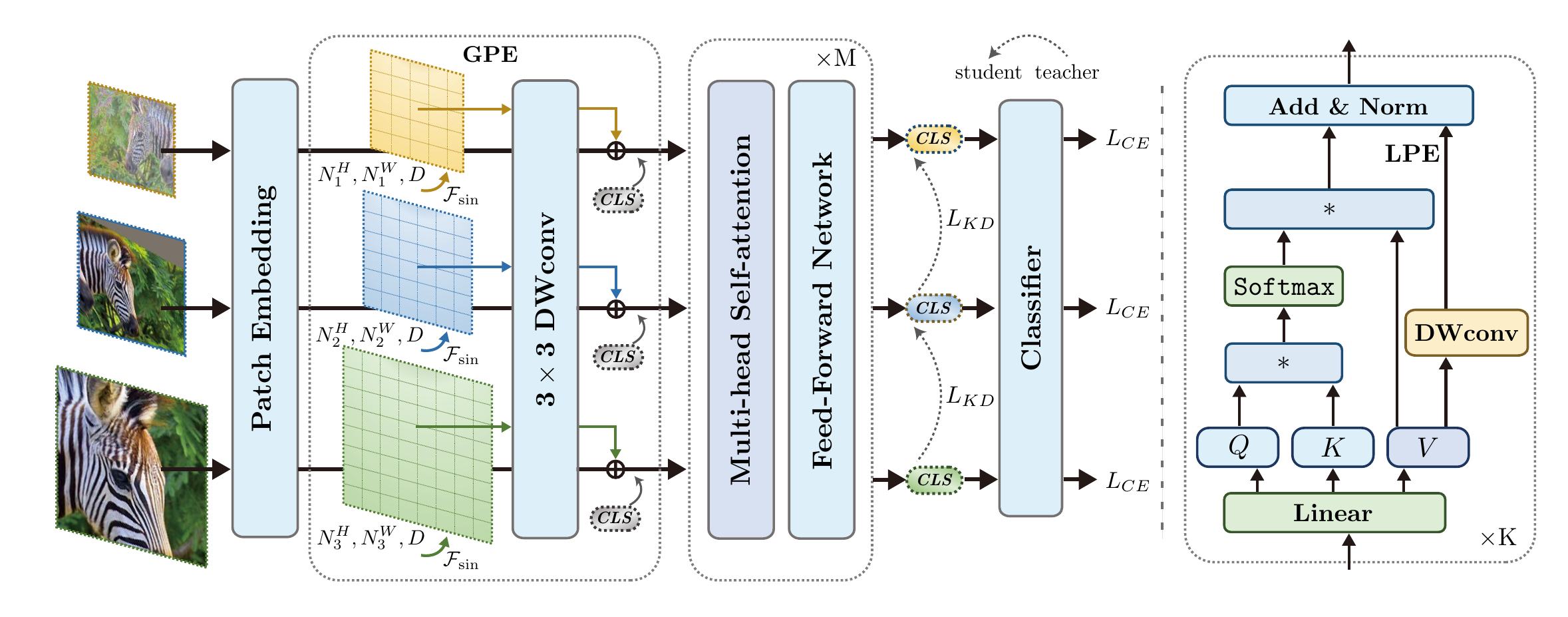}
\vspace{-0.2in}
\caption{\textbf{Left}: The overview of \system framework. \textbf{Right}: The pipeline of generating local positional embedding.}
\label{fig:network}
\vspace{-0.05in}
\end{figure*}

\section{Method} 

Our goal is to train a vision transformer that not only performs well on resolutions the network has seen during training, but more importantly it is able to adapt to a wide range of unseen resolutions without significant performance drop during testing. To this end, we first introduce a resolution scaling transformer in \cref{sec:res_method}, \system, which operates on input samples of multiple resolutions in the training stage. 
Since the size of objects varies in different scales, we also introduce a scale consistency loss to fully explore information from all resolutions for improved  accuracy. Furthermore, as mentioned in \cref{sec:explore_res}, directly interpolating positional embeddings to unseen resolutions during inference produces unsatisfactory results. To mitigate this issue, \system builds upon carefully designed global-local positional embeddings, which are generated conditioned on input resolutions, as will be described in \cref{sec:pe}.

\subsection{\system}
\label{sec:res_method}
Following the vanilla ViT~\cite{dosovitskiy2020image}, given an input image $X$ whose height and width are $H$ and $W$, respectively, we first split it into $N^H \times N^W$ patches, where the patch size is set to $t$ and $N_H = H / t, N_W = W / t$. Each image patch is projected into a $D$-dimension feature by patch embedding and is denoted as a ``token''. Subsequently, a global class token \texttt{cls} is concatenated with image tokens before they are fed into Transformer blocks.
 
Unlike standard ViTs operating on single-scale images, \system takes inputs of different resolutions during training so as to better model objects of varying sizes in different scales and generalize better during inference. More specifically, as shown in \cref{fig:network}, we replicate a given training image for $r$ times, where $r$ denotes the number of resolutions used. For the $i$-th data replica, resizing and cropping operations\footnote{In practice, we realize it with \texttt{RandomResizedCrop} in PyTorch.} are applied to obtain a training sample ${X^i}$ with a spatial size of $3\times H_i \times W_i$. Afterwards, we apply random pre-processing strategies involved in ViTs training paradigm\footnote{Random pre-processing includes Auto-Augment~\cite{cubuk2018autoaugment}, RandAugment~\cite{cubuk2020randaugment}, random erasing~\cite{zhong2020random}, MixUp~\cite{zhang2017mixup} and CutMix~\cite{yun2019cutmix}.} on each scale of inputs separately. As a result, one mini-batch is composed of groups of multi-resolution inputs sharing identical labels, which is roughly equivalent to extending the base batch size by $r$ times. In addition, for the input sample ${X^i}$, we feed the global class token output of the last transformer block as inputs into classification head to compute final predictions $Y^i$. Naturally, the classification losses can be written as:
\vspace{-0.01in}
\begin{equation}
    \label{eqn:ce}
    L(\Theta) =   \underset{({X}, \, T) \thicksim \mathcal{D}} {\mathbb{E}}    \sum_{i=1}^{r} L_{CE}(Y^i; X^i, T, \Theta),
\end{equation}
where $T$ denotes the ground-truth label and $L_{CE}$ represents the cross-entropy loss. In addition, $\mathcal{D}$ and $\Theta$ denote the training set and the parameters of the network, respectively.

\vspace{0.05in}
\noindent \textbf{Scale consistency loss.} Given that larger inputs generally produce better recognition results compared to their smaller counterparts, we take advantage of knowledge distillation through enforcing consistencies among different resolutions. In particular, we use a smooth $l_1$ loss with feature whitening~\cite{wei2022contrastive}, denoted as $L_{KD}$, to transfer knowledge from the class token of a higher resolution to that of a lower resolution. This is achieved by serving $\texttt{cls}_{i}$ as the teacher of $\texttt{cls}_{i+1}$ with $H_{i} = W_i, H_{i} > H_{i + 1}$. Combining with~\cref{eqn:ce}, the loss can be written as:
\vspace{-0.01in}
\begin{align}
\label{eqn:kd}
L(\Theta) =   \underset{({X}, \, T) \thicksim \mathcal{D}} {\mathbb{E}} \,\,  \frac{1}{r}[ & \sum_{i=1}^{r} L_{CE}(Y^i; X^i, T, \Theta) \nonumber \\
 +  & \sum_{i=1}^{r-1} L_{KD}(\texttt{cls}_{i+1},\texttt{cls}_{i})].
\end{align}
Especially, teacher class tokens are detached from the gradient computational graph. At last, the loss is divided by $r$ to ensure stability of training.

\subsection{Global-Local Positional Embedding} 

\label{sec:pe}

The commonly-used positional embeddings highly depend on the size of input samples. As a result, when multiple resolutions are involved in the training process, positional embeddings need to be carefully adjusted when iterating images of different scales, as simple interpolations incur performance drops. Therefore, we propose to use conditional positional embeddings both globally and locally to bridge the resolution gap among a broad range of resolutions. Below, we first introduce the global positional embedding and then describe its local counterpart.

\vspace{0.05in}
\noindent \textbf{Global position embedding.} To incorporate location information in patch embeddings, a typical way is to add absolute position embedding (APE). Given the input sample $X$,  $x^{\text{img}}$ refers to the output image tokens of the patch embedding, whose spatial dimension equals $N_H \times N_W$ and feature dimension is $D$. For simplicity, we denote $x$ as concatenation of class token \texttt{cls} and image tokens $x^{\text{img}}$, the absolute positional embedding $p$ can be expressed as, 
\begin{equation}
x = x + p, \quad p \in \mathbb{R}^{1 \times (N^H \times N^W + 1) \times D}.
\end{equation}

The most straightforward way is implementing $p$ with learned parameters, as widely-adopted in~\cite{touvron2021training,dosovitskiy2020image,wang2021pyramid}. Another common tactic is fulfilled with sinusoidal mapping $\mathcal{F}_{sine}$~\cite{vaswani2017attention, he2022masked}, through which $p$ is generated on-the-fly by a fixed function dependent on $N_H, N_W $ and $D$. 
Due to space limitation, we provide explicit expression in \cref{sec:appendix_image}. 
Furthermore, compared with learned APE, the sine-cosine APE changes more weakly between different input scales, as displayed in \cref{sec:ablate_res} . Therefore, we build our method upon sine-cosine APE with the assumption that smoother positional embedding would contribute to better resolution scalability. 

We further improve the sine-cosine positional embedding with conditional computation such that the embeddings are tailored to the model during training. As illustrated in \cref{fig:network}, a simple yet effective depth-wise convolution is applied so as to generate the final positional embedding conditioned on sinusoidal encoding. Since convolutions should be performed in 2-D dimension, we leave out the class token by concatenating a zero padding shaped of $\mathbb{R}^{1 \times 1\times D}$ with output embeddings of DWconv. In general, the strategy introduced above aims at injecting smooth spatial information of global context into ViT, thereby we denote it as global positional embedding (GPE).

\vspace{0.05in}
\noindent \textbf{Local positional embedding.} Positional embeddings introduced in~\cite{zhang2021self,dong2022cswin,li2022uniformer} share the same design philosophy since they are both dynamically generated by input tokens and carry spatial information of local neighbourhood. It has been unveiled that such strategies can effectively introduce translation invariance into ViTs and hence facilitate generalizing to various resolutions. We refer to the positional embedding conditioned on local input feature as local positional embedding (LPE) and hypothesize that LPE is orthogonal with GPE in modelling spatial information of image tokens. Consequently, the combination of LPE and GPE may results in best resolution scalability.

To this end, we incorporate local positional embeddings into attention blocks in a similar fashion to~\cite{dong2022cswin}. Given a multi-head self-attention block,  a query $Q$, a key $K$ and a value $V$ are obtained through a linear projection, and the output $z$ can be derived as:
\begin{equation}
\label{eq:attn_old}
z = \mathtt{Softmax}(Q K^T / \sqrt{D})V.
\end{equation}

\vspace{-0.05in}
In particular, local spatial information of $V$ is utilized. We first set the class token aside and reshape the value matrix to get $V' \in \mathbb{R}^{M \times D' \times N_H \times N_W}$, where $M$ denotes the number of attention heads and $D'$ satisfies $D = D' \cdot M $. Inspired by~\cite{liu2021swin, liu2022swin}, we generate dynamic positional embeddings conditioned on $V'$ separately for each head. Therefore, a $3\times 3$ depth-wise convolution is implemented to obtain the LPE for each head. The above operations can be denoted as mapping $\mathcal{H}$ conditioned on $V$. Therefore, \cref{eq:attn_old} can be re-written as:
\begin{equation}
z = \mathtt{Softmax}(Q K^T / \sqrt{D})V + \mathcal{H}(V).
\end{equation}

\vspace{-0.05in}
By virtue of convolutions, LPE can be dynamically generated regardless of input scales. Eventually, in \system, global and local positional embeddings are combined to ensure better generalization to novel resolutions.

\begin{table*}[t]
  \centering
  \caption{Top-1 Accuracy of DeiT and \system on ImageNet-1K. Columns highlighted with \colorbox{Gray}{grey background} refer to the training resolutions of given models. Specifically, \system adopts training resolutions of 128, 160 and 224 for multi-resolution training.}
  \small
  \addtolength{\tabcolsep}{2.5pt}
  
  \ra{1.1}
{\begin{tabular}{l|ccccccccccc}
        \toprule
        \multirow{2}{*}{\textbf{Model}} & \multicolumn{11}{c}{\textbf{Testing resolution}} \\

        ~ &  96 & 112 & 128 & 160 & 192 & 224  & 288 & 384 & 448 & 512 & 640\\
        \cmidrule{1-2} \cmidrule{3-12} 
        DeiT-T~\cite{touvron2021training} & 8.06& 34.22 & 52.16& 65.68& 70.18& \colorbox{Gray}{72.14}& 73.1& 71.29& 67.43& 66.07& 59.31\\
        \system-T-MR & \textbf{61.40} & \textbf{64.93} & \colorbox{Gray}{\textbf{67.78}} & \colorbox{Gray}{\textbf{71.09}} & \textbf{72.97} & \colorbox{Gray}{\textbf{73.85}} & \textbf{74.85} & \textbf{75.04} & \textbf{74.39} & \textbf{73.77} & \textbf{71.65}\\
        \midrule
        DeiT-S~\cite{touvron2021training}  & 17.55 & 54.34 & 67.02 & 75.62 & 78.60 &\colorbox{Gray}{79.83} &  80.02 & 78.10 & 75.85 & 72.63 & 63.86\\
        \system-S-128 & 70.25& 73.91 & \colorbox{Gray}{75.47} & 77.06 & 77.48 & 76.89 & 74.78		 & 69.55 & 64.54 & 58.34 & 45.25\\
        \system-S-160 & 67.34& 72.26 & 75.05& \colorbox{Gray}{78.06}& 78.94& 79.19& 78.25& 	74.86& 71.38& 66.65& 54.77\\
        \system-S-224 & 57.80 &66.36 & 71.35 & 76.99 & 79.63 &  \colorbox{Gray}{80.83} & 81.42 & 80.65&79.28&77.73&73.26\\
         \system-S-MR  &  \textbf{73.59}& \textbf{76.64} & \colorbox{Gray}{\textbf{78.24}} & \colorbox{Gray}{\textbf{80.39}} & \textbf{81.42} & \colorbox{Gray}{\textbf{82.16}} & \textbf{82.70} & \textbf{82.72} & \textbf{82.52} &\textbf{82.00} & \textbf{80.72}\\
        \midrule
        DeiT-B~\cite{touvron2021training} & 27.86 & 64.46 & 73.18 & 79.05 & 81.06 & \colorbox{Gray}{81.79} & 82.19 & 81.11 & 79.81& 78.23& 74.23\\
        \system-B-MR  & \textbf{75.86}& \textbf{78.42} & \colorbox{Gray}{\textbf{79.74}} & \colorbox{Gray}{\textbf{81.52}} & \textbf{82.28} & \colorbox{Gray}{\textbf{82.72}} & \textbf{83.02} & \textbf{83.29} & \textbf{82.9} & \textbf{82.63} & \textbf{81.72}\\
    \bottomrule
    \end{tabular}}
  \vspace{-0.15in}
\label{tab:main_res}
\end{table*}

\section{Experiments}

\vspace{0.05in}
\noindent \textbf{Implementation details.} We instantiate \system with DeiT~\cite{touvron2021training} due to its simplicity. Given an input image, we resize it to 128, 160 and 224, respectively, for multi-resolution training throughout the experiments, unless specified otherwise. The resulting images are then used as inputs to \system. For image classification, we use AdamW~\cite{kingma2014adam} as our optimizer and apply a cosine decay learning rate scheduler. Small and tiny models are trained with a batch size of 1024 and a learning rate of $5e^{-4}$, yet a learning rate of $8e^{-4}$ is used for the base model. We keep all augmentation and regularization settings in~\cite{touvron2021training} for fair comparisons. For all experiments, we follow the official training and testing split as well as the evaluation metrics.
More details can be found in \cref{sec:appendix_image}. 
For testing, we report results on a wide range of resolutions. Note that \system only uses a single scale during testing.

\subsection{Main Results}
\noindent \textbf{Effectiveness of \system in image classification.}
\cref{tab:main_res} presents the results of \system and comparisons with DeiT~\cite{touvron2021training} using various settings. In particular, we use \system-M-R to denote a variant of \system, where M represents the model size (\ie, T, S, B for tiny, small and base models respectively) and R indicates the resolution used for training (\ie, MR denotes multiple resolution; if R is a number, it represents the resolution itself). 

When evaluated with a testing resolution of 224, {\system}s achieve highly competitive results---\system-S-MR and \system-B-MR  offers an accuracy of 82.16\% and 82.72\%, respectively, outperforming their DeiT counterparts by 2.33\% and 0.93\%. We also see from \cref{tab:main_res} that \system trained with multi-resolution images outperforms models trained with single scale inputs with clear margins on all ``seen'' resolutions. For instance, \system-S-MR outperforms \system-S-128,  \system-S-160, \system-S-224 by 2.77\%, 2.33\% and 1.33\% respectively.  Similar trends can also be found for \system-B and \system-T. This highlights the effectiveness of multi-resolution training.

\begin{table}[t]
  \centering
  \caption{Results and comparisons of different backbones on ADE20K. All backbones are pre-trained on ImageNet-1k, among which MAE~\cite{he2022masked} uses unsupervised pre-training. }
    \vspace{-0.1in}
    \small
    \addtolength{\tabcolsep}{-2.0pt}
     \ra{1.1}
    {\begin{tabular}{l|cccc}
        \toprule
        {\textbf{Backbone}} & {\textbf{\#Param}} & {\textbf{Lr schd}}& {\textbf{mIoU}} & {\textbf{ms + flip}}\\
        \cmidrule{1-2} \cmidrule{3-5} 
        DeiT-S~\cite{touvron2021training} & 52.1M & 80k & 42.96 & 43.79\\
        XCiT-S12/16~\cite{ali2021xcit} & 52.4M& 160k & 45.90 & 46.72\\
        \system-S-224 & 51.7M &  80k & 45.47 & 46.61\\
        \system-S-MR & 51.7M & 80k & \textbf{46.31} & \textbf{47.45}\\
        \midrule
        DeiT-B~\cite{touvron2021training} & 120.6M & 160k & 45.36& 47.16 \\
        XCiT-S24/16~\cite{ali2021xcit} & 109.0M & 160k & 47.69 & 48.57\\
        ViT-B + MAE~\cite{he2022masked} & 176.5M & 160k & 48.13 & 48.70\\
        \system-B-MR & 119.8M& 160k & \textbf{48.30} & \textbf{49.28}\\
    \bottomrule
    \end{tabular}}
\vspace{-0.1in}
\label{tab:ade20k_res}
\end{table}

Furthermore, for ``unseen'' resolutions, \system demonstrates clear scaling capabilities. In particular, given a test resolution of 384, \system-S-224  achieves a Top-1 accuracy of 80.65\%, which is 2.55\% higher than its DeiT-S counterpart (78.10\%). This suggests that global-local positional embeddings can indeed improve generalization of different resolutions. \system-S-MR further boosts the accuracy to 82.72\%, demonstrating the benefit of multi-resolution training. Besides, \system consistently generalize well to lower resolutions. Compared with DeiT, \system-S-MR and \system-B-MR increase by 56.24\% and 48.00\% when evaluated on a resolution of 80, highlighting the effectiveness of \system when dealing with significant resolution shifts during inference.

\begin{table}[t]
    \centering
    \small
    \addtolength{\tabcolsep}{-4.5pt}
    \caption{Results and comparisons of different backbones on the mini-val set of COCO2017 using Mask R-CNN~\cite{he2017mask} and 3$\times$ training schedule. All backbones are pre-trained on ImageNet-1k in the supervised setting. Part of results are credited to~\cite{ali2021xcit,chen2022vision}.}
    \vspace{-0.1in}
    \ra{1.1}
    {\begin{tabular}{l|cccccccc}
    \toprule
    \textbf{Backbone} & \textbf{\#Param} & \textbf{AP$^{b}$}  & \textbf{AP$^{b}_{50}$} & \textbf{AP$^{b}_{75}$}& \textbf{AP$^{m}$} & \textbf{AP$^{m}_{50}$} & \textbf{AP$^{m}_{75}$}  \\
    \midrule
    PVT-Small~\cite{wang2021pyramid} & 44.1M & 43.0 & 65.3& 46.9& 39.9&62.5 &42.8 \\
    XCiT-S12/16~\cite{ali2021xcit} & 44.3M & 45.3 & 67.0& 49.5& 40.8&64.0 & 43.8\\
    ViT-S~\cite{li2021benchmarking} & 43.8M & 44.0 & 66.9 & 47.8 & 39.9& 63.4& 42.2\\
    ViTDet-S~\cite{li2022exploring}& 45.7M & 44.5 & 66.9& 48.4& 40.1& 63.6& 42.5\\
    \system-S-MR & 45.6M & \textbf{46.4} & 68.5& 50.4& \textbf{40.7} & 64.7 & 43.4\\
    \midrule
    PVT-Large~\cite{wang2021pyramid} & 81.0M & 44.5 & 66.0 & 48.3& 40.7&63.4&43.7 \\
    XCiT-M24/16~\cite{ali2021xcit} & 101.1M & 46.7 & 68.2& 51.1& \textbf{42.0}& 65.6 & 44.9 \\
    ViT-B~\cite{li2021benchmarking} & 113.6M & 45.8 &68.2 & 50.1& 41.3 & 65.1& 44.4\\
    ViTDet-B~\cite{li2022exploring}&121.3M & 46.3& 68.6& 50.5& 41.6& 65.3 & 44.5\\
    \system-B-MR & 115.3M  & \textbf{47.6}& 69.0 & 52.0 & 41.9 & 65.9 & 44.4 \\
    \bottomrule
    \end{tabular}}
    \vspace{-0.05in}
    \label{tab:res_od}
\end{table}

\vspace{0.05in}
\noindent \textbf{Semantic segmentation.} To show flexibility of \system, We evaluate  
 for semantic segmentation on ADE20K~\cite{zhou2019semantic} with UperNet~\cite{xiao2018unified}. 
 As shown in \cref{tab:ade20k_res}, \system-S-224 improves DeiT-S by 2.51 measured by 
  mIoU. Both \system-S-MR and \system-B-MR, which are pre-trained with the multi-resolution strategy, achieve better results. In particular, \system-S-MR reaches up to 47.45 and \system-B-MR hits the peak of 49.28 mIoU. This suggests that \system effectively models multi-scale and high-resolution features for pixel-level dense predictions. Note that \system-B-MR and \system-S-MR are directly used as pre-trained backbones and we do not perform multi-resolution fine-tuning on ADE20K, since segmentation tasks already require images with a size of $512\times512$ as inputs, and multi-resolution training would be computationally expensive. Nonetheless, results in \cref{tab:ade20k_res} demonstrate the great potential of transferring models that are pre-trained with multiple resolutions for dense prediction tasks.

\vspace{0.05in}
\noindent \textbf{Object detection.}
We further explore performance of \system on COCO2017~\cite{lin2014microsoft} for object detection and instance segmentation, following the designs of ViTDet~\cite{li2022exploring} by appending simple feature pyramids on the feature maps of last-layer outputs and using both non-shifted window attention and global self-attention blocks. In addition, To adapt from \system pre-trained on ImageNet-1K, we also adopt global positional embedding and inject local positional embeddings into all attention blocks. According to results reported in \cref{tab:res_od}, \system achieves promising results, \eg \system-S-MR outperforms ViTDet-S by 2.0 box AP and 0.6 mask AP and \system-B-MR surpasses ViTDet-B by 1.3 box AP and 0.3 mask AP. We believe that the improved resolution scalability of \system contributes to better performance on object detection.

\begin{table}[t]
  \centering
  \small
    \addtolength{\tabcolsep}{-2.0pt}
  \caption{Top-1 Accuracy of TimeSformer on Kinetics-400. MR stands for multi-resolution training.}
    \vspace{-0.1in}
{\ra{1.1}
\begin{tabular}{l|ccccc}
        \toprule
        \multirow{2}{*}{\textbf{Model}}  &  \multicolumn{5}{c}{\textbf{Testing resolution}} \\
        &  96 &   128 & 160 & 224 & 288\\
        \cmidrule{1-3} \cmidrule{4-6} 
        TimeSFormer~\cite{bertasius2021space} & 26.28 & 61.94&  70.60  &\colorbox{Gray}{75.54} &75.45 \\
         \system-B-224  & 58.61 & 68.50 & 73.09 & \colorbox{Gray}{76.32} & 76.78 \\
         \system-B-160  & 67.28 & 71.64& \colorbox{Gray}{74.56} & 75.98 & 75.18\\
         \system-B-128  &  64.66 & \colorbox{Gray}{72.32} & 74.13 & 74.19 & 72.51\\
        \system-B-MR  & 70.56 & \colorbox{Gray}{74.33}& \colorbox{Gray}{76.38} & \colorbox{Gray}{77.32} & 77.56 \\
    \bottomrule
    \end{tabular}}
\label{tab:k400_res}
\vspace{-0.1in}
\end{table}

\vspace{0.05in}
\noindent \textbf{Video action recognition.} We also evaluate \system for video action recognition on Kinetics400. For an easy adaption from our pre-trained image models to the video domain, we choose the TimeSFormer~\cite{bertasius2021space} framework with a divided spatial and temporal attention design. In particular, we initialize the backbone with weights of model pre-trained on ImageNet-1K and conduct multi-resolution training on Kinetics400 with clip sizes set to $8 \times 224\times 224$, $8 \times 160\times 160$ and $8 \times 128\times 128$ respectively. As \cref{tab:k400_res} demonstrates, \system-B fine-tuned with single resolution outweighs vanilla TimeSFormer by 0.78\% on a testing resolution of 224 and generalizes better to clips of both higher and lower resolutions. On top of that, by implementing multi-resolution training on video samples, \system improves performance on each training resolution by a large margin, \eg, the Top-1 accuracy on testing resolution of 160 grows from 73.09\% to 76.08\%.

\subsection{Discussion} \label{sec:ablate_res} 

\begin{figure}[t]
  \centering
  \includegraphics[width=0.95\linewidth]{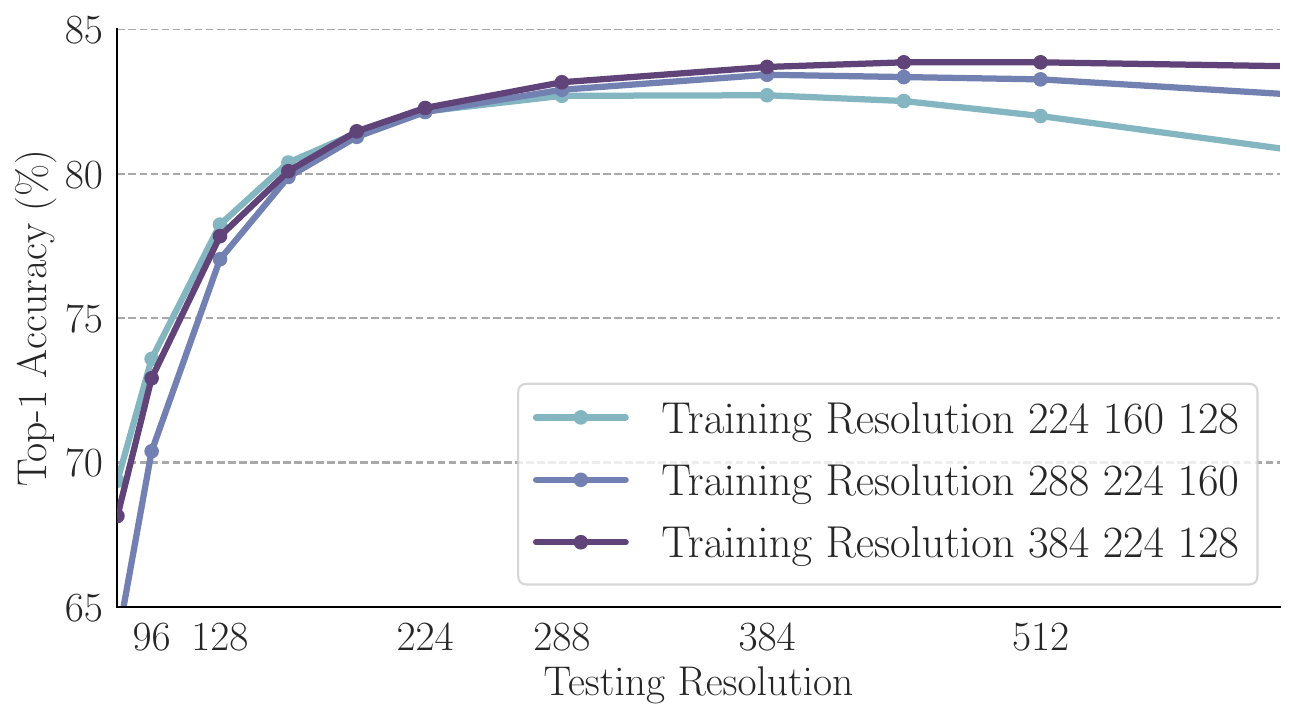}
    \vspace{-0.1in}
   \caption{Top-1 Accuracy of \system-S-MR with different training resolutions on ImageNet-1K.}
   \label{fig:res_ablation}
\end{figure}

\noindent \textbf{Training resolutions.} We experiment with 3 different  settings using a small model, \ie, (128, 160, 224), (160, 224, 288) and (128, 224, 384), which we denoted as (a), (b) and (c) respectively.  The results are summarized in \cref{fig:res_ablation}. We see that \system achieves outstanding performance on a wide range of resolutions. More specifically, compared with (a), (b) adopts higher training resolutions, consequently reflecting on performance rise in high-resolution inputs and incurring a drop on low resolution. Regarding differences between setting of (b) and (c), the spectrum of training resolutions expands in both directions. Despite the wide range between 128 and 384, we witness a all-around improvement of (c) over (b), highlighting that \system is able to deal with significant resolution variations.

\begin{figure}[t]
  \centering
  \includegraphics[width=0.95\linewidth]{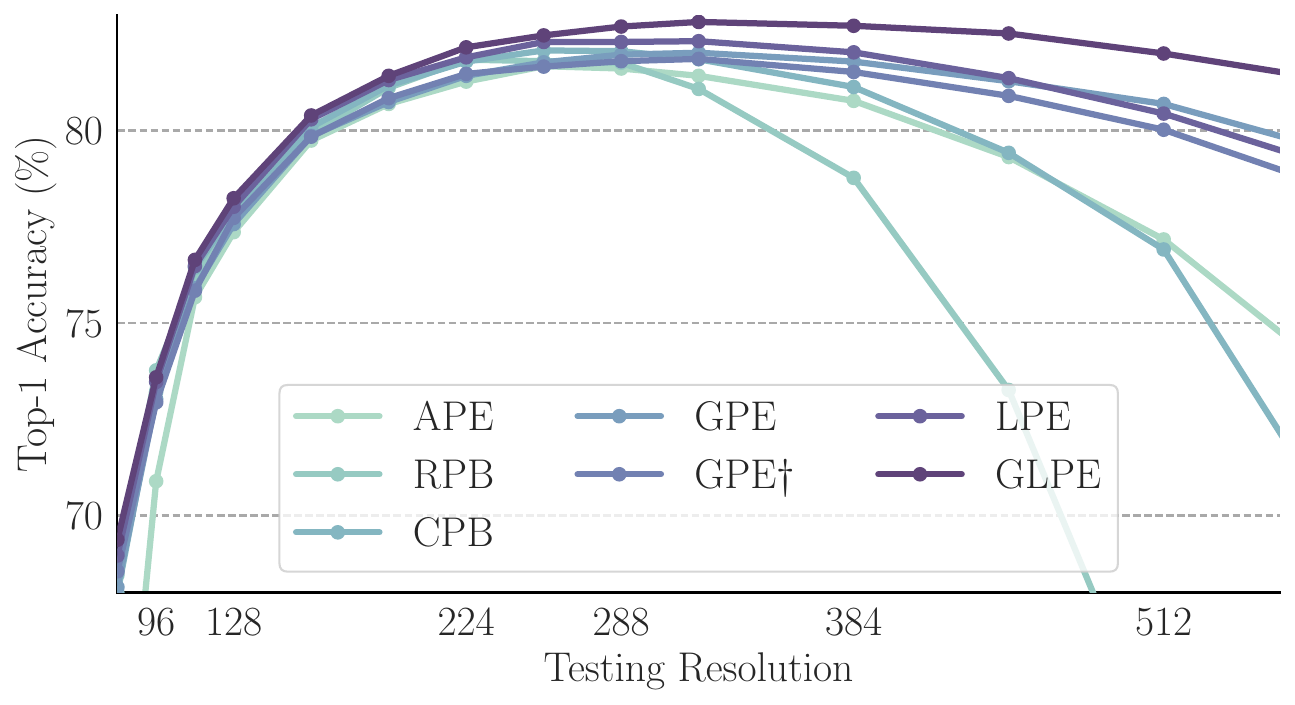}
    \vspace{-0.1in}
   \caption{Results of different positional embedding strategies on a broader range of resolutions.}
   \label{fig:pe_ablation}
    \vspace{-0.1in}
\end{figure}

\vspace{0.05in}
\noindent \textbf{Positional embedding.} We evaluate the performance of \system with different positional embedding strategies using a small model. In particular, we compare with 
\begin{enumerate*}[label=(\arabic*)]
\item APE, which stands for vanilla absolute positional embedding in DeiT~\cite{touvron2021training}. In practice, we set the spatial dimension of APE according to the highest training resolution and downsample it for lower resolutions. For inference, the position embedding can be re-scaled to any test resolution with bicubic interpolation;
\item APE* which uses an individual APE for each training resolution;
\item RPB, which is introduced in~\cite{liu2021swin} and we use the RPB of highest resolution for interpolation during inference;
\item CPB, which is a resolution-agnostic strategy~\cite{liu2022swin} and images of arbitrary scales can be input into ViTs with CPB directly;
\item GPE, which is our global positional embedding;
\item GPE$\dagger$, which represents the plain sine-cosine absolute positional embedding without convolutional enhancement;
\item LPE, which is our local positional embedding;
\item GLPE, which is the combination our GPE and LPE.
\end{enumerate*}

\cref{tab:ablation_pe} shows the results of different positional embeddings on training resolutions.
We see that interpolating APE makes no differences compared with maintaining multiple APEs (\ie, APE*), which suggests that ViTs can be trained to deal with different scales of inputs with shared spatial information. Furthermore,
all positional embeddings coupled with multi-resolution training demonstrate better results compared to their counterparts trained with single resolutions, \ie, steady gains are made by all positional embeddings when testing on 128, 160 and 224, (gains are shown in the bracket in \cref{tab:ablation_pe}).
\cref{fig:pe_ablation} further presents the results of generalizing to more resolutions. Clear performance drops can also be observed in \cref{fig:pe_ablation} when APE, RPB and CPB are scaled up to unseen large resolutions, especially RPB. In contrast, LPE and GPE decreases slowly towards extremely large resolutions. GLPE, the combination of LPE and GPE, offers the best results.

\begin{table}[t]
  \centering
  \caption{Results of \system-S-MR with different positional embedding (PE) strategies on ImageNet-1K. The performance gain compared to single-resolution training is indicated in the bracket.}
    \vspace{-0.1in}
      \ra{1.1}
      \small
    {\begin{tabular}{l|ccc}
        \toprule
        \multirow{2}{*}{\textbf{PE}}  & \multicolumn{3}{c}{\textbf{Testing resolution}} \\
         &  128 & 160 & 224\\
        \cmidrule{1-1} \cmidrule{2-4}
        APE & 77.36~\rise{3.99} & 79.74~\rise{2.46} & 81.27~\rise{1.44}\\
        APE* &  77.31~\rise{3.93} &  79.58~\rise{2.21} & 81.42~\rise{1.59}\\
        RPB & 77.90~\rise{2.74} & 79.92~\rise{2.04} & 81.84~\rise{1.27}\\
        CPB & 77.64~\rise{1.33} & 79.77~\rise{1.77} & 81.74~\rise{1.13}\\
        GPE$\dagger$ & 77.73~\rise{2.73} & 79.83~\rise{2.27} & 81.47~\rise{1.29}\\
        GPE & 77.57~\rise{2.76}& 79.63~\rise{2.05}& 81.42~\rise{1.40} \\
        LPE & 78.02~\rise{2.62}& 80.29~\rise{2.29}& 81.90~\rise{1.28}\\
        GLPE& \textbf{78.24~\rise{2.77}}& \textbf{80.39~\rise{2.33}}& \textbf{82.16~\rise{1.33}}\\
    \bottomrule
    \end{tabular}}
    \vspace{-0.1in}
\label{tab:ablation_pe}
\end{table}

\vspace{0.05in}
\noindent \textbf{Knowledge distillation.} To strengthen the interaction between different resolutions, we use a smooth-L1 loss to distill information from class tokens. We also experiment with a L2 loss (\ie, Mean squared error). Further, as inputs of different resolutions output features with different scales, we additionally follow the practice in DeiT~\cite{touvron2021training} by distilling logits with a Kullback-Leibler divergence loss.
The experiments are conducted on \system-S-MR for 100 epochs for efficiency purposes.  \cref{tab:ablation_kd} shows the ablation results. We observe the efficacy of distilling with class tokens compared to logits. In addition, the smooth $L_1$ loss have similar performance with $L_2$ loss with slightly better results on high resolutions (\ie, 224).

\begin{table}[t]
  \centering
  \caption{Results of \system-S-MR with different distillation strategies on ImageNet-1K for 100ep. Performance gains over result of training \textbf{without distillation} are shown in the bracket.}
    \vspace{-0.1in}
    \small
    \addtolength{\tabcolsep}{-2.0pt}
    {\begin{tabular}{cc|ccc}
        \toprule
        \multicolumn{2}{c|}{\textbf{Distillation}}  & \multicolumn{3}{c}{\textbf{Testing resolution}} \\
        Target & Loss &128 & 160 & 224\\
        \cmidrule{1-2} \cmidrule{3-5} 
         logit & KL & 73.50~\equal{0.0} & 76.45~\rise{0.07} & 78.82~\rise{0.26}\\
         cls & $L_2$ & 74.71~\rise{1.21} & 77.27~\rise{0.89} & 79.33~\rise{0.77}\\
         cls & smooth $L_1$ & \textbf{74.71~\rise{1.21}} & \textbf{77.39~\rise{1.01}}& \textbf{79.68~\rise{1.12}}\\
    \bottomrule
    \end{tabular}}
    \vspace{-0.1in}
\label{tab:ablation_kd}
\end{table}

\vspace{0.05in}
\noindent \textbf{Training strategies.} We also explore a widely-used multi-resolution training strategy~\cite{he2016deep,he2017mask} without cross-scale consistency loss, where one iteration consists of randomly sampled images of one certain resolution. In particular, we feed samples of different scales (\ie, 128, 160, 224) iteratively 
based on two settings: (1) iteration-based, where each mini-batch uses one resolution and resolutions vary for different training iterations; (2) epoch-based, where a fixed resolution is used for each epoch and the change of resolutions only occur at the epoch-level. As \cref{tab:ms_ablation} shows, both iteration-based and epoch-based multi-resolution training generate worse results compared to single resolution training. In contrast, our strategy demonstrates strong advantages on all training resolutions by clear margins, even without the scale-consistency loss, highlighting the importance of enforcing consistencies of all resolutions in a mini-batch.  

\begin{table}[t]
  \centering
  \caption{Results of \system-S-MR with different training strategies on ImageNet-1K. We append performance gain/drop compared with single-resolution training in the bracket.}
  \vspace{-0.1in}
    \small
    {\begin{tabular}{l|ccc}
        \toprule
        {\textbf{Training}} & \multicolumn{3}{c}{\textbf{Testing resolution}} \\
        {\textbf{Strategy}} &128 & 160 & 224\\
        \cmidrule{1-1} \cmidrule{2-4}
        MR (iter) & 75.70~\rise{0.23} & 78.31~\rise{0.25} & 80.32~\drop{0.51}\\
        MR (epoch) & 75.18~\drop{0.29} & 78.05~\drop{0.01} & 80.26~\rise{0.16} \\
        MR w/o KD & 77.72~\rise{2.25} & 79.66~\rise{1.60} & 81.77~\rise{0.94}\\
        MR & \textbf{78.24}~\rise{2.77}& \textbf{80.39}~\rise{2.33}& \textbf{82.16}~\rise{1.33} \\
    \bottomrule
    \end{tabular}}
  \vspace{-0.05in}
\label{tab:ms_ablation}
\end{table}

\begin{figure}[h]
\centering
\vspace{-0.05in}
\includegraphics[width=\linewidth]{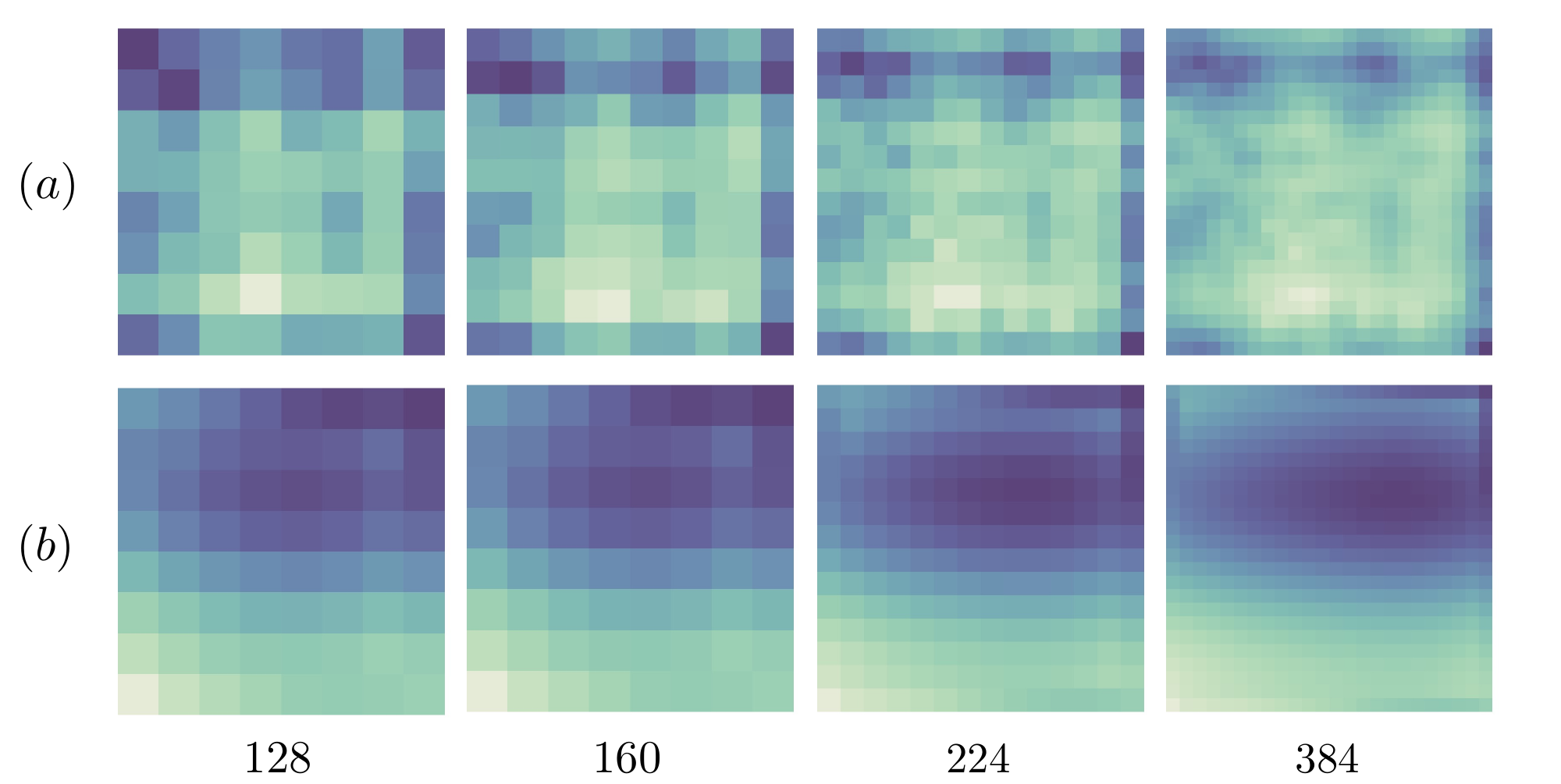}
\vspace{-0.2in}
\caption{Heatmaps of different PE averaged on each token. \textbf{(a)}: Absolute Positional Embeddings (APE),  \textbf{(b)}: Global Positional Embeddings (GPE). }
\label{fig:pe_vis}
\vspace{-0.2in}
\end{figure}

\vspace{0.05in}
\noindent \textbf{Qualitative visualizations.} We visualize two positional embeddings on resolutions of  128, 160, 224 and 384, respectively. As shown in \cref{fig:pe_vis}, compared with APEs shifted by interpolation, our GPEs that are generated with convolutions demonstrate a smoother variations among input scales. In addition, \cref{fig:pe_ablation} suggests that GPE generalizes better to higher resolutions unseen in training.

\section{Conclusion}
We introduced \system, a ViT framework to encourage excellent all-round performance on a wide range of resolutions. In particular, \system was motivated by training on sample of different scales and aided by a scale-consistency loss. A global-local positional embedding strategy was also introduced to facilitate better generalization on unseen resolutions. Extensive experiments demonstrated promising scalabilities of \system in a broad range of resolutions. We also observe that \system can be readily adapted to downstream tasks, \eg, semantic segmentation, object detection and video action recognition.

\vspace{0.05in}
\textbf{Acknowledgement} This project was supported by NSFC under Grant No. 62102092 and No. 62032006.

{\small
\bibliographystyle{ieee_fullname}
\bibliography{egbib}
}

\appendix
\section{More Experiments}

\subsection{More about Resolution Scalability}

\begin{figure}[b]
  \centering
  \includegraphics[width=0.95\linewidth]{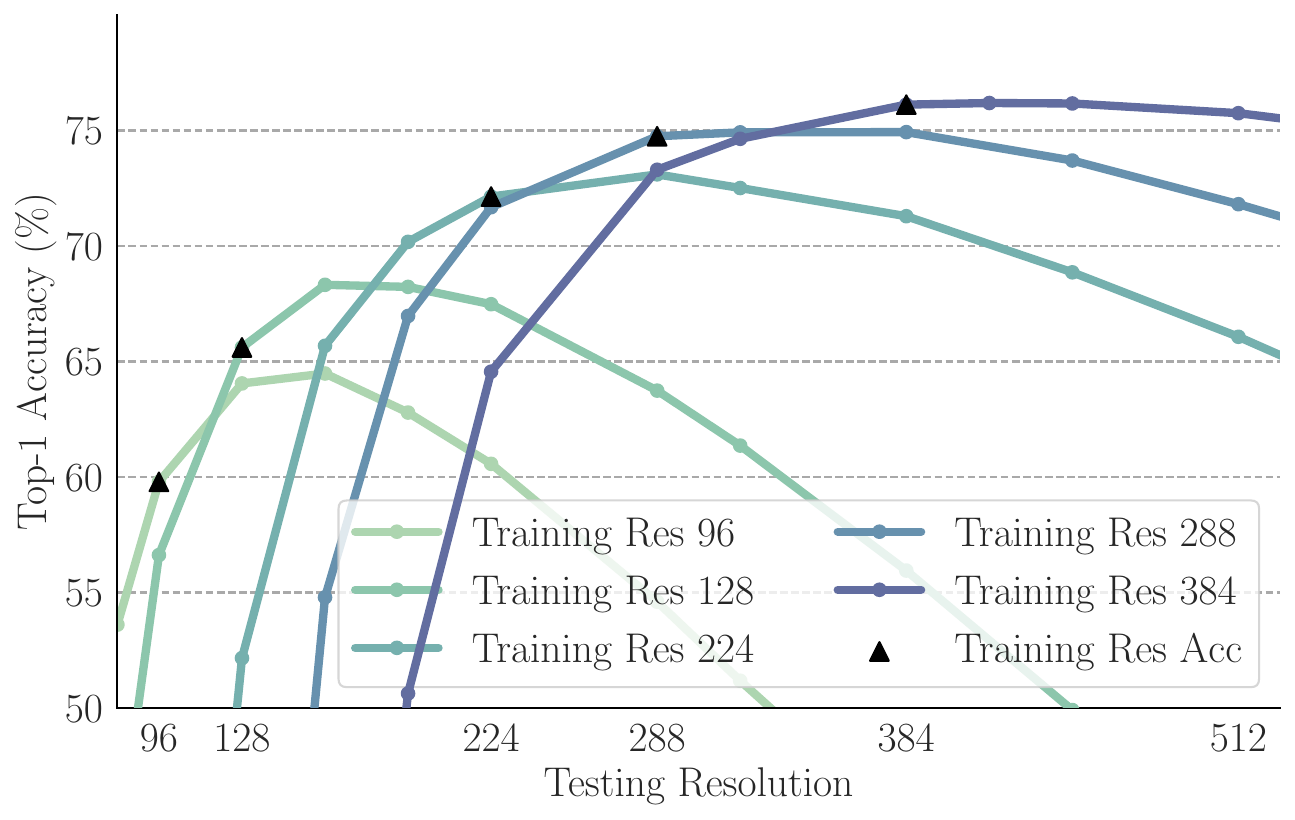}
    \vspace{-0.1in}
   \caption{Top-1 accuracy of DeiT-T trained with 5 different resolutions and tested on resolutions varying from 80 to 576.}
   \label{fig:res_tiny_test}
    \vspace{-0.1in}
\end{figure}

\begin{figure}[b]
  \centering
  \includegraphics[width=0.95\linewidth]{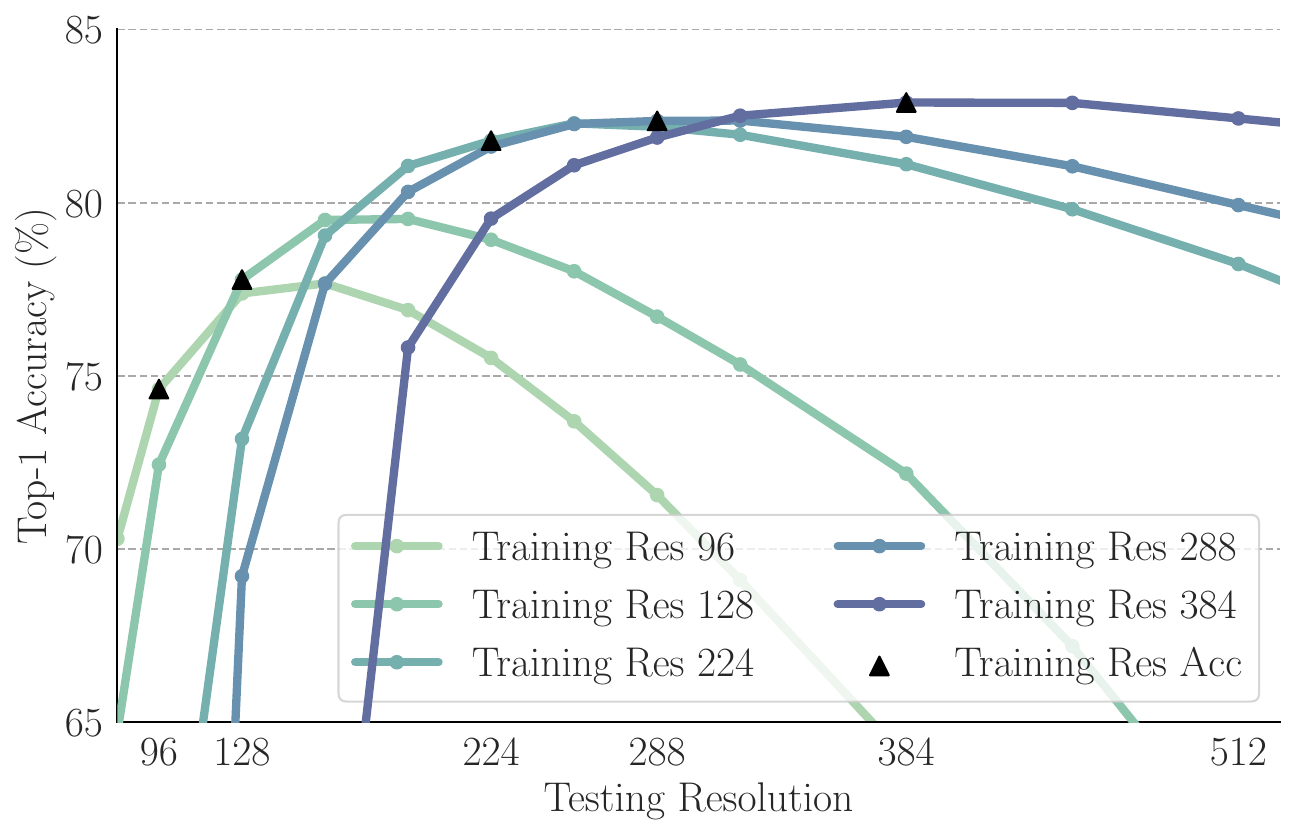}
    \vspace{-0.1in}
   \caption{Top-1 accuracy of DeiT-B trained with 5 different resolutions and tested on resolutions varying from 80 to 576.}
   \label{fig:res_base_test}
    \vspace{-0.1in}
\end{figure}

\noindent \textbf{Scalability of vanilla ViTs.} As displayed in \cref{fig:res_tiny_test} and \cref{fig:res_base_test}, in order to provide more comprehensive insights into resolution scalability, we further test tiny and base models of DeiT~\cite{touvron2021training} which are pre-trained on training resolutions of 196,128, 224, 288 and 384, respectively. The evaluation is conducted by generalizing models to different testing resolutions ranging from 80 to 576. We can observe that the trends towards scaling up and scaling down testing resolutions are consistent with ones on DeiT-S.

\noindent \textbf{Extending the range of testing resolutions.} To further explore the potential for \system, we extend the range of testing resolutions to  1024. As shown in \cref{tab:scale_up}, compared with DeiT, \system achieves much more decent performance on fairly testing resolutions. 
\begin{table}[t]
  \centering
    \caption{Comparison of Top-1 accuracy between DeiT and \system on ImageNet-1K with high testing resolutions.}
    \small
    \addtolength{\tabcolsep}{-0.5pt}
    {\begin{tabular}{l|ccccc}
        \toprule
        \multirow{2}{*}{\textbf{Model}} & \multicolumn{4}{c}{\textbf{Testing resolution}} \\
         & 512 & 640 & 800 & 1024\\
        \cmidrule{1-2}  \cmidrule{3-5} 
        DeiT-S-224 & 72.63 & 63.86 & 49.31& 31.45 \\
        \system-S-MR (224) & 82.00 & 80.72& 78.12& 72.49\\
        \cmidrule{1-2}  \cmidrule{3-5} 
        DeiT-S-384 & 81.09 &79.35 &75.67 &67.61\\
        \system-S-MR (384) & 83.86 & 83.71 & 83.37 & 82.58 \\
    \bottomrule
    \end{tabular}}
\label{tab:scale_up}
\vspace{-0.05in}
\end{table}

\subsection{Evaluation on Robustness Datasets}
We also evaluate our models on ImageNet-related robustness datasets, \ie, ImageNet-Rendition (IN-R)~\cite{hendrycks2021many}, ImageNet-A (IN-A)~\cite{hendrycks2021nae}, ImageNet-Sketch (IN-SK)~\cite{wang2019learning}, ImageNet-C (IN-C)~\cite{hendrycks2019robustness} and ImageNetv2 (IN-v2)~\cite{recht2019imagenet}. As reported in \cref{tab:res_robust}, we observe that \system achieves promising performance on robustness as well. For example, \system-S-224 is superior to DeiT-S on each dataset while \system-S-MR makes further improvements. In particular, on IN-A, \system-S-MR surpasses \system-S-224 by 7.88 \% and DeiT-S by 10.07\%. This suggests that training with multi-scale inputs facilitates ViTs to cope with hard as well as out-of-distribution inputs.

\begin{table}[htbp]
    \centering
    \caption{Performance on ImageNet-based robustness benchmarks. mCE~\cite{hendrycks2019robustness} is employed for IN-C while Top-1 accuracy is used for IN-R, IN-A and IN-SK.}
    \small
    \addtolength{\tabcolsep}{-2.0pt}
    {\begin{tabular}{l|ccccc}
    \toprule
    \textbf{Model} & \textbf{IN-R}$\uparrow$ & \textbf{IN-A}$\uparrow$ & \textbf{IN-SK}$\uparrow$ & \textbf{IN-C}$\downarrow$ & \textbf{INv2}$\uparrow$ \\
    \midrule
    DeiT-S~\cite{touvron2021training} & 41.93& 19.84& 29.09 &  54.60 & 68.47\\
    \system-S-224 & 43.95 & 22.03 & 30.91 & 52.31 & 69.81\\
    \system-S-MR & 45.08 & 29.91 & 31.47 & 51.03 & 71.68\\
    \midrule 
    DeiT-B~\cite{touvron2021training} &44.66& 28.15& 31.96& 48.52 & 70.91\\
    \system-B-MR & 45.38 & 33.89 & 33.06 & 48.83 & 71.88\\

    \bottomrule
    \end{tabular}}
    \label{tab:res_robust}
\end{table}

\subsection{Training Efficiency}

\begin{figure}[b]
  \centering
  \includegraphics[width=0.95\linewidth]{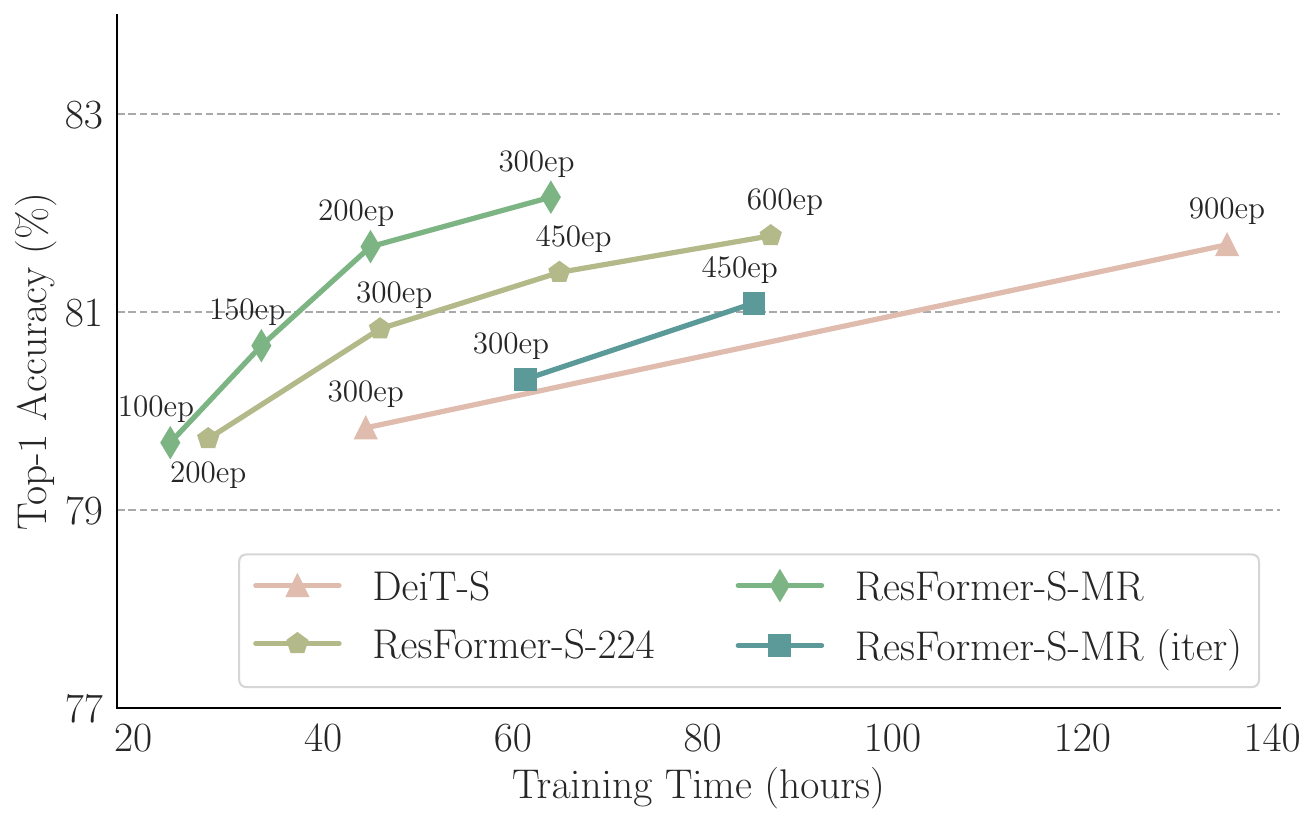}
    \vspace{-0.1in}
   \caption{Trade-off between training time and Top-1 Accuracy on ImageNet-1K with a testing resolution of 224. Same hardware and software settings are adopted for all experiments, \ie, we utilize 8$\times$ V100-32GB GPUs and set the per-GPU batch size to 128.}
   \label{fig:eff_map}
    \vspace{-0.1in}
\end{figure}

\begin{figure}[b]
  \centering
  \includegraphics[width=0.95\linewidth]{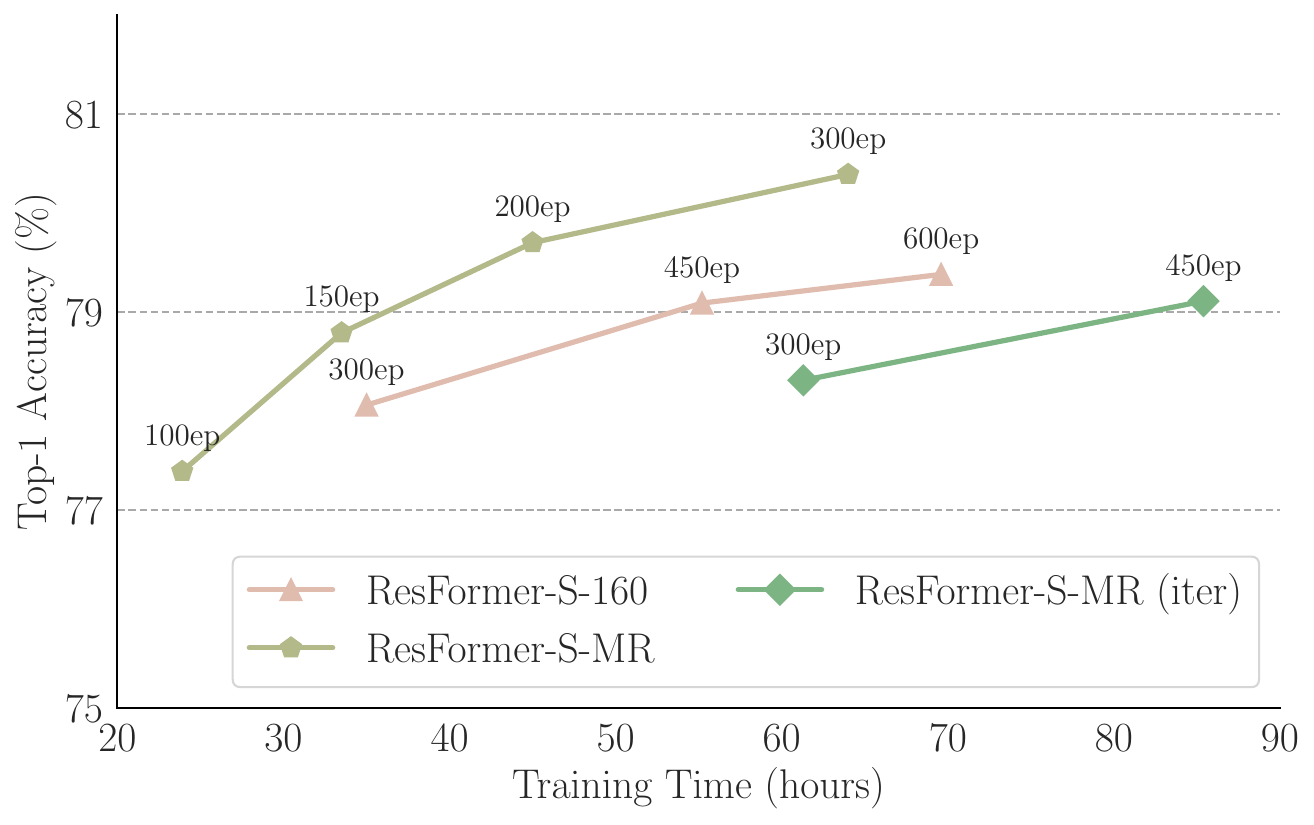}
    \vspace{-0.1in}
   \caption{Trade-off between training time and Top-1 Accuracy on ImageNet-1K with a testing resolution of 160. Same hardware and software settings are adopted for all experiments.}
   \label{fig:eff_map_160}
    \vspace{-0.1in}
\end{figure}

During the training of \system, each input sample is replicated by $r$ times, and thus this increases the training time. For efficiency, we reduce the total number of training epochs to 200, 150 and 100 respectively while keeping other hyperparameters unchanged. As shown in \cref{fig:eff_map}, \system demonstrates competitive performance on training efficiency. For instance, \system-S-MR with 200-epoch training surpasses the 300-epoch counterparts \system-S-224 and DeiT-S by 0.83\% and 1.83\% in Top-1 accuracy despite that they share similar training time.

As depicted in \cref{fig:eff_map_160}, training with a single lower resolution (\ie, 160) significantly saves time. Nevertheless, \system-S-MR still has an edge on time-performance trade-off, \eg, \system-S-160 with 450-epoch training is more time-consuming than \system-S-MR with 200-epoch training while the accuracy is 0.61\% lower.

\section{Implementation Details}
\subsection{Image Classification}
\label{sec:appendix_image}
\vspace{0.05in}
\noindent \textbf{Sine-Cosine positional embedding.} We demonstrate the explicit mapping function $\mathcal{F}_{sine}$ for sine-cosine positional embedding $p$ as follows. Firstly, image tokens are placed in a 2D spatial dimension as $x^{img} \in \mathbb{R}^{N_H \times N_W \times D}$. We denote the  positional embedding for the token coordinated at $(m, n)$ as $p_{m,n} \in \mathbb{R}^{1\times D}$. Particularly, $d$-th dimension of $p_{m,n}$ can be mapped with $\mathcal{F}_{sine}(m, n, d)$ as below,

\begin{equation*}
\begin{aligned}
\mathcal{F}_{sin}(m,n,d) &= 
\begin{cases}
\mathit{f}_{sin}(m, d , N_H, D) &\quad \text{if } d < D/2 \\
\mathit{f}_{sin}(n, d , N_W, D) &\quad \text{otherwise} \\
\end{cases}, \\
\mathit{f}_{sin}(pos, d, N, D) &= 
\begin{cases}
\sin (\frac{pos}{N + \epsilon} / T^{2d/D} )  &\text{if } d\%2=0 \\
\cos (\frac{pos}{N + \epsilon} / T^{2(d-1)/D})  & \text{otherwise}\\
\end{cases}, \\
\end{aligned}
\vspace{0.05in}
\end{equation*}
where the temperature $T$ and  $\epsilon$ is set to 10000 and $1e^{-6}$ respectively, and a normalization is also used to ensure better continuity among varying resolutions. For simplicity, $N_i^H, N_i^W, D$ are omitted from function parameters.

\vspace{0.05in}
\noindent \textbf{Detailed hyperparameters.} For experiments of image classification on ImageNet-1K, we set the hyperparameters for training \system-T, \system-S, \system-B froom scratch and fine-tuning on DeiT according to \cref{tab:param_class}. 

\vspace{0.05in}
\noindent \textbf{Augmentation strategy.} Motivated by unsupervised learning, we apply separate random augmentation on different scales of inputs. In particular, to ensure the consistency of class tokens between different scales, as an exception, we apply MixUp~\cite{zhang2017mixup} and CutMix~\cite{yun2019cutmix} across different scales with same variables. As shown in \cref{tab:aug_ablate}, separate augmentation slightly outperform its counterpart, especially on the lowest testing resolution.

\begin{table}[h]
  \centering
    \vspace{-0.02in}
  \caption{Ablation study of augmentation strategies.}
    \vspace{-0.05in}
{
\small
\begin{tabular}{lc |ccc}
        \toprule
        \multirow{2}{*}{\textbf{Model}} & \textbf{Sep} & \multicolumn{3}{c}{\textbf{Testing resolution}} \\
         & \textbf{Aug} & 128 & 160 & 224\\
        \cmidrule{1-2}  \cmidrule{3-5} 
        \system-S-MR & & 77.50 & 80.14 & 81.93\\
        \system-S-MR & \checkmark &  78.24 & 80.39 & 82.16 \\
    \bottomrule
    \end{tabular}}
\label{tab:aug_ablate}
\vspace{-0.1in}
\end{table}

\begin{table}[t]
    \centering
    \caption{Hyperparameters for training on ImageNet-1K.}
    \begin{tabular}{l|p{1.2cm}<{\centering} p{1.2cm}<{\centering} p{1.2cm}<{\centering}}
    \toprule
    \multirow{2}{*}{\textbf{Hyperparameters}} &\textbf{Tiny /} & \multirow{2}{*}{\textbf{Base}} & \textbf{Fine-}\\
    ~ & \textbf{Small} & ~ & \textbf{tune}\\
    \toprule
    Epochs & 300 & 200 & 30 \\
    Base learning rate & 5e-4 & 8e-4 & 5e-5 \\ 
    Warmup epochs  &  5& 20 & 5 \\
    Stoch. depth & 0.1 & 0.2& 0.1 \\
    Gradient clipping & \xmark & 5.0 & \xmark \\
    \midrule
    Batch size & \multicolumn{3}{c}{1024} \\
    Weight decay & \multicolumn{3}{c}{0.05} \\
    Optimizer & \multicolumn{3}{c}{AdamW} \\
    Learning rate schedule & \multicolumn{3}{c}{Cosine} \\    
    \midrule
    Repeated augmentation & \multicolumn{3}{c}{\checkmark} \\
    Random erasing  & \multicolumn{3}{c}{0.25} \\
    Random augmentation  & \multicolumn{3}{c}{9/0.5} \\
    Mixup  & \multicolumn{3}{c}{0.8}     \\
    Cutmix   & \multicolumn{3}{c}{1.0}    \\
    Color jitter & \multicolumn{3}{c}{0.4} \\

    \bottomrule
    \end{tabular}
    \label{tab:param_class}
\end{table}

\subsection{Semantic Segmentation}
We follow the common practice on ADE20K~\cite{zhou2019semantic} by training on $512 \times 512$ inputs for 80k iterations for \system-S and for 160k iterations for \system-B, respectively. In addition, we employ the AdamW optimizer with a learning rate of $6e^{-5}$, a weight decay of 0.01 and a batch-size of 16. We base our implementation on MMSegmentation~\cite{mmseg2020} and adopt the corresponding augmentations, \ie, random resizing with the ratio range set to (0.5, 2.0), random horizontal flipping with probability of 0.5 and random photometric distortion. Despite that \system employs a columnar structure, we simply extract features from different layers (\ie the $2$nd, $5$th, $8$th and $11$th layers) as inputs of UperNet~\cite{xiao2018unified} without FPN-like necks. We report results in two different testing settings. For the first one, inputs are scaled to having a shorter side of 512. In addition, we apply flipping on inputs of multiple scales that are varied in (0.5, 0.75, 1.0, 1.25, 1.5 1.75) $\times$ of training resolutions.

\subsection{Object Detection} To further validate the efficacy of \system on dense prediction tasks, we evaluate \system on COCO 2017~\cite{lin2014microsoft} for object detection and instance segmentation. In particular, we adopt Mask R-CNN~\cite{he2017mask} as our framework based on MMDetection~\cite{chen2019mmdetection} and train with the 3$\times$ schedule. Furthermore, we utilize AdamW optimizer with a learning rate of $1e^{-4}$, weight decay of 0.05 and a batch size of 16. It is worth noting that we follow the common multi-scale training for object detection instead of fine-tuning with the multi-resolution strategy. Therefore, training samples are resized randomly so that the shorter sizes vary from 480 to 800 with step of 32 and the longer sides are within 1333.

\subsection{Video Action Recognition}
Similar to the implementation for images, we train \system  on videos by replicating video clips to get multi-scale copies. Specifically, given a certain sampling rate $s$ of $1/32$, a clip $X$ of $F = 8$ frames is sampled and replicated into $r$ copies. Different cropping sizes are applied on each sequence of frames. Consequently the $i$-th training copy $X_i$ is sized in $\mathbb{R}^{F \times H_i \times W_i}, i \in \{1, \cdots, r\}$. We also keep the augmentation strategy used for images by applying separate random augmentations~\cite{bertasius2021space} on each clip.

We follow the divided attention design adopted in TimeSFormer~\cite{bertasius2021space}, in which attention computation is conducted along spatial dimension and temporal dimension separately. In order to align with image models, we only incorporate global and local positional embeddings into into spatial dimensions.  For training on Kinetics-400~\cite{kay2017kinetics}, we adopt the same strategy with TimeSFormer~\cite{bertasius2021space}. In particular, the training epoch is set to 15 and the initial learning rate is set to $5e^{-3}$. In addition, we employ a SGD optimizer and a multi-step scheduler which divides the learning rate by 10 times at the 11th and the 14th epoch respectively.

In particular, we observe that \system achieves better performance on videos with $L_2$ scale consistency loss. In order to improve performance by ensuring coherence in pre-training and fine-tuning. We adapt \system-B-MR for $L_2$ loss for an extended fine-tuning of 100 epochs which matches the common 300-epoch pre-training. For fair comparison, we initiate all \system{s} in Kinetics-400 downstream tasks with same pre-trained weights.

\end{document}